\pdfoutput=1
\documentclass[10pt]{article}

\usepackage{arxiv}
\usepackage[utf8]{inputenc}
\usepackage[T1]{fontenc}
\usepackage{microtype}

\usepackage{
    amsmath,
    amsfonts,
    amssymb,
    amsthm,
    mathtools,
    bm,
    bbm,
    nicefrac
}
\usepackage{thm-restate}

\usepackage{graphicx}
\graphicspath{{./images/}}

\usepackage{booktabs}
\usepackage{array}
\usepackage{multirow}

\usepackage{algorithm}
\usepackage{algpseudocode}

\usepackage{enumitem}

\usepackage[numbers,sort&compress]{natbib}

\usepackage{xcolor}

\usepackage[most]{tcolorbox}
\newtcolorbox{highlightbox}{
    colback=yellow!20,
    colframe=yellow!20,
    boxrule=0pt,
    arc=0pt,
    left=4pt,
    right=4pt,
    top=4pt,
    bottom=4pt
}

\usepackage[
    colorlinks=true,
    linkcolor=blue,
    citecolor=blue,
    urlcolor=blue
]{hyperref}

\newcommand{\R}{\ensuremath{\mathbb{R}}}

\DeclareMathOperator*{\argmax}{arg\,max}

\newcommand{\vtheta}{\bm{\theta}}

\newcommand{\vveps}{\bm{\varepsilon}}

\newcommand{\ve}{\bm{e}}

\newcommand{\vg}{\bm{g}}

\newcommand{\vv}{\bm{v}}

\newcommand{\vx}{\bm{x}}

\newcommand{\mA}{\bm{A}}
\newcommand{\mB}{\bm{B}}

\newcommand{\mE}{\bm{E}}

\newcommand{\mG}{\bm{G}}
\newcommand{\mH}{\bm{H}}

\newcommand{\mM}{\bm{M}}

\newcommand{\mU}{\bm{U}}
\newcommand{\mV}{\bm{V}}
\newcommand{\mW}{\bm{W}}

\newcommand{\mSigma}{\bm{\Sigma}}

\renewcommand{\shorttitle}{SAM and Muon: Robustness under the Spectral Norm}

\title{%
Sharpness-Aware Minimization and Muon:\\
Robustness under the Spectral Norm
}

\author{%
  Wenzhi Zhong \\
  Department of Mathematical Sciences \\
  University of Bath \\
  \texttt{wz912@bath.ac.uk}
  \And
  Edward Milsom \\
  Department of Computer Science \\
  University of Bath \\
  \texttt{em846@bath.ac.uk}
  \And
  Michael Murray \\
  Department of Mathematical Sciences \\
  University of Bath \\
  \texttt{mjm253@bath.ac.uk}
}

\date{}

\begin{document}

\maketitle

\begin{abstract}
    Sharpness-Aware Minimization (SAM) aims to improve generalization by encouraging insensitivity to small, worst-case parameter perturbations. However, the notion of a “small” perturbation is inherently geometry-dependent: while existing SAM variants have explored a wide range of choices,
    a clear perspective on which geometries are most effective in practice remains elusive. 
    Recent work on matrix-aware optimization, particularly the Muon optimizer, suggests that respecting the matrix structure of hidden-layer weights can lead to strong empirical performance. Motivated by this, we study matrix-aware geometry in both stages of SAM: we introduce a layerwise spectral inner perturbation for matrix-valued hidden-layer parameters and combine it with either AdamW/SGDW or Muon in the outer update. Across ImageNet-1K experiments on ViT-Small/16 and ResNet-50, we find that the combination of a spectral inner step with a Muon outer step performs consistently strongly, achieving the best validation accuracy on both models among the evaluated methods. 
\end{abstract}

\section{Introduction}
The flat minimum hypothesis proposes that neural network parameter settings in flat regions of the loss landscape yield better generalizing networks \citep{flat-minima}. Inspired by this, Sharpness-Aware Minimization \citep[SAM;][]{foret2020SAM} is an optimization algorithm which encourages solutions in flat basins by minimizing the loss subject to a perturbation of the weights, therefore seeking neighbourhoods that have uniformly low loss. Concretely, the SAM algorithm uses a two-step procedure: we first perturb the weights in the uphill direction of the gradient (the \textit{inner step}), and then compute the descent direction from that perturbed point (the \textit{outer step}), applying the actual step to the original unperturbed parameters. This original formulation of SAM defines sharpness in terms of Euclidean geometry, i.e.,
the inner perturbation step is taken within a fixed Euclidean norm radius, while the outer descent step traditionally uses SGD, whose update direction is the negative Euclidean gradient.

Recently, a separate line of work has proposed that steepest descent under a spectral norm step-size penalty could be more appropriate for neural network optimization, since it takes into account the linear operator structure of weight matrices, instead of just treating them as flattened vectors of parameters \citep{bernstein2025modularduality}. This has led to a rise in popularity of \textit{matrix-aware} optimizers, in particular Muon \citep{jordan2024muon}, which is now an integral component of many frontier large language model training setups \citep{kimiteam2025kimik2openagentic,5team2025glm45agenticreasoningcoding,singh2026arceetrinitylargetechnical,deepseek2026v4}. While various works have extended SAM to incorporate non-Euclidean geometry \citep[such as Fisher, Riemannian, and adaptive geometries,][]{pmlr-v162-kim22f,yun2023riemanniansam,pmlr-v139-kwon21b}, to our knowledge, no work has studied the effects of spectral geometry in either the inner or outer step of the SAM setting.

In this work, we investigate the use of spectral geometry in the inner and/or outer step. In particular, in the outer step we consider a Muon update, while in the inner step we consider an orthogonalized gradient perturbation\footnote{This is similar to a  Muon ascent step, but without momentum.}. We test these modifications on two ImageNet training tasks (a vision transformer and a ResNet) and find the following:

\begin{itemize}[itemsep=0pt,topsep=0pt]
\item Inner and outer geometry interact. Spectral sharpness helps most when the outer update is also spectral and least when it is Euclidean, whereas Euclidean sharpness is comparatively insensitive to the outer optimizer.
\item Spectral perturbations with Muon achieve the best validation accuracy on both ViT-S/16  and ResNet-50.
\end{itemize}

\section{Preliminaries}
\subsection{A norm-agnostic template for SAM}\label{subsec:geometric-sam}
The flat minimum hypothesis suggests we should judge a parameter choice not only by its training loss, but also by its training loss \emph{under small perturbations}. Following~\cite{foret2020SAM} we focus on robustness under worst-case perturbations: let $\Theta \cong \mathbb{R}^p$ denote the parameter space, $\|\cdot\|_{\mathrm{in}}$ be a norm on $\Theta$, and $L:\Theta\to\mathbb{R}_{\ge 0}$ be the training loss. For perturbation size $\rho\ge 0$ we define the \emph{robust loss} associated with $L$ as
\begin{equation}\label{eq:robust-loss}
    L_\rho(\vtheta) = \sup_{\|\vveps\|_{\mathrm{in}}\le 1} L(\vtheta+\rho \vveps).
\end{equation}
Our goal here is not to give an implementation-level description of SAM, involving batching etc., but rather to isolate a general first-order template common to SAM-type methods which allows their comparison. Since evaluating Eq.~\eqref{eq:robust-loss} is generally intractable, the approach taken in~\cite{foret2020SAM} is to replace the inner maximization by its first-order linearization. In what follows, we formulate this linearization using the Fréchet differential, rather than immediately identifying it with a gradient vector. We do this in order to present a more general SAM framework in which the perturbation geometry can be determined by a possibly non-Euclidean norm \(\|\cdot\|_{\mathrm{in}}\)\footnote{If \(\Theta \cong \mathbb{R}^p\) is equipped with its standard Euclidean inner product, \(DL(\vtheta)\) is typically represented by \(\nabla L(\vtheta)\)}. Assume \(L\) is Fréchet differentiable at \(\vtheta\). We write \(DL(\vtheta)\in \Theta^*\) (where $\Theta^*$ is the dual space of $\Theta$) for its differential, which recall in this case is the linear functional satisfying
\[
    L(\vtheta+\vveps)
    =
    L(\vtheta) + DL(\vtheta)[\vveps] + o(\|\vveps\|_{\mathrm{in}})
    \qquad \text{as } \vveps\to 0 .
\]
Thus, for small \(\rho\),
\[
    L(\vtheta+\rho\vveps)
    =
    L(\vtheta)
    +
    \rho\,DL(\vtheta)[\vveps]
    +
    o(\rho),
    \qquad \|\vveps\|_{\mathrm{in}}\le 1 .
\]
The first-order adversarial perturbation is any steepest-ascent
direction for the linear functional \(DL(\vtheta)\) over the
\(\|\cdot\|_{\mathrm{in}}\)-unit ball,
\begin{equation}\label{eq:opt-inner-pert}
    \vveps^*(\vtheta)
    \in
    \argmax_{\|\vveps\|_{\mathrm{in}}\le 1}
    DL(\vtheta)[\vveps].
\end{equation}
Equivalently, using the canonical identification \(\Theta \cong \Theta^{**}\), $\vveps^*(\vtheta)$ is an element of the subdifferential of the dual norm, $\vveps^*(\vtheta) \in \partial \|DL(\vtheta)\|_{\mathrm{in},*}$.
From the definition of the dual norm it therefore follows that
\footnote{Note also by Fréchet differentiability the remainder is uniform over the
\(\|\cdot\|_{\mathrm{in}}\)-unit ball}
\begin{equation}\label{eq:robustlossdualinnernormpenalty}
    L_\rho(\vtheta)
    =
    L(\vtheta)
    +
    \rho\,\|DL(\vtheta)\|_{\mathrm{in},*}
    +
    o(\rho).
\end{equation}
This first-order calculation motivates the following SAM surrogate for the robust loss $L_{\rho}$, in which one simply evaluates the loss at the first-order adversarially perturbed point,
\begin{equation}\label{eq:perturbed-surrogate}
\vtheta\mapsto L(\vtheta+\rho\vveps^*(\vtheta)).
\end{equation}
In particular, the surrogate given in Eq.~\eqref{eq:perturbed-surrogate} agrees with $L_{\rho}$ up to first order in \(\rho\) and is convenient to work with. If one is able to differentiate this surrogate exactly, then the dependence of \(\vveps^*(\vtheta)\) on \(\vtheta\) contributes an additional term:
if \(L\) is differentiable at \(\vtheta+\rho\vveps^*(\vtheta)\) and a
differentiable selection of \(\vveps^*\) exists, then the chain rule gives
\[
    D\big(L(\vtheta+\rho\vveps^*(\vtheta))\big)[\Delta]
    =
    DL(\vtheta+\rho\vveps^*(\vtheta))
    \big[
        \Delta+\rho\,D\vveps^*(\vtheta)[\Delta]
    \big].
\]
However, in general, the maximizer map \(\vveps^*\) may be non-unique or
non-differentiable, as is the case for non-smooth norms (such as the spectral norm we consider shortly), and even
when a differentiable selection exists, the term
\(D\vveps^*(\vtheta)\) involves second-order information about \(L\). This is typically too expensive to compute in large-scale training. Standard SAM therefore uses a stop-gradient approximation: it forms the perturbation \(\vveps^*(\vtheta)\), evaluates the loss at
\(\vtheta+\rho\vveps^*(\vtheta)\), and uses only the differential of \(L\) at that perturbed point. We denote this stopped-gradient first-order signal by
\[
    \phi_\rho(\vtheta)
    :=
    DL(\vtheta+\rho\vveps^*(\vtheta))
    \in \Theta^* .
\]
At this point, the SAM perturbation mechanism has produced a covector
\(\phi_\rho(\vtheta)\in\Theta^*\) (i.e. a perturbed-loss gradient). An outer algorithm (i.e. an optimizer) must turn this into a parameter-space update direction. We represent this operation by a possibly time and state dependent direction map
\[
    \mathcal{D}_t:\Theta^*\to\Theta,
    \qquad
    \Delta_t = \mathcal{D}_t(\phi_\rho(\vtheta_t)).
\]
\begin{algorithm}[t]
\caption{Abstract SAM Template}
\label{alg:general-SAM}
\begin{algorithmic}[1]
\Require Initial parameters $\vtheta_0$; inner norm $\|\cdot\|_{\mathrm{in}}$;
perturbation radii $(\rho_t)_{t\ge 0}$; step sizes $(\eta_t)_{t\ge 0}$;
outer direction maps $(\mathcal{D}_t)_{t\ge 0}$.
\For{$t=0,1,\dots,T-1$}
    \State Compute an inner steepest ascent direction:
    $\vveps_t \in \partial\|DL(\vtheta_t)\|_{\mathrm{in},*}$
    \State Form the perturbed first-order signal:
    $ \phi_t := DL(\vtheta_t+\rho_t\vveps_t)$.
    \State Convert the perturbed signal into an update direction:
    $ \Delta_t = \mathcal{D}_t(\phi_t)$. 
    \State Update the parameters: $\vtheta_{t+1} = \vtheta_t+\eta_t\Delta_t$.
\EndFor
\State \textbf{return} $\vtheta_T$
\end{algorithmic}
\end{algorithm}
The abstraction of SAM given in Algorithm~\ref{alg:general-SAM} separates two ways in which SAM-type methods can vary. First, they may differ in the geometry used to define the inner adversarial perturbation: classical SAM~\citep{foret2020SAM} uses the Euclidean norm, ASAM~\citep{pmlr-v139-kwon21b} uses a parameter-dependent perturbation geometry which may be written as $\|\vveps\|_{\mathrm{in},\vtheta} = \|T_{\vtheta}^{-1}\vveps\|_2$ with \(T_{\vtheta}\) diagonal, and Fisher / Functional SAM~\citep{pmlr-v162-kim22f,pmlr-v267-singh25b} use a norm defined through the model Jacobian so that perturbations are measured by their functional effect on the network output. Second, SAM-type methods may differ in how they convert the perturbed
first-order signal $\phi_t$ into an update direction. For vanilla SGD,
\(\mathcal{D}_t\) is essentially the negative Euclidean gradient direction, up to scaling and mini-batch noise. Momentum, Adam~\citep{2015-kingma}, AdamW~\citep{loshchilov2018decoupled}, and related optimizers instead apply state-dependent transformations to the same perturbed signal.

\subsection{Orthogonalized matrix updates and Muon}\label{subsec:muon}
A central operation in what follows is the orthogonalization of a matrix-valued update. Let $\mM \in \R^{m\times n}$ and write its reduced singular value decomposition as $
\mM = \mU_r \mSigma_r \mV_r^\top$.
We define its orthogonalization by
\[
\text{Ortho}(\mM) := \mU_r \mV_r^\top.
\]
Thus $\text{Ortho}(\mM)$ preserves the left and right singular subspaces of $\mM$ while replacing all non-zero singular values by $1$. This operation admits a natural geometric interpretation: by the duality between the spectral and nuclear norms~\citep{bhatia97},
\begin{equation}\label{eq:ortho-characterization}
\text{Ortho}(\mM)\in
\argmax_{\|\Delta\|_{2\to 2}\le 1}
\langle \mM,\Delta\rangle,
\end{equation}
where $\|\Delta \|_{2\to 2}$ denotes the spectral norm of $\Delta$, i.e. its maximum singular value.
Thus \(\text{Ortho}(\mM)\) is a steepest-ascent direction for the linear
functional \(\Delta\mapsto\langle \mM,\Delta\rangle\) under a spectral-norm
constraint. An orthogonalized update changes the action of the layer by at most
\(\eta\) in spectral norm:
\[
    \|(\mW-\eta\operatorname{Ortho}(\mM))-\mW\|_{2\to2}
    =
    \eta\|\operatorname{Ortho}(\mM)\|_{2\to2}
    \le \eta.
\]
Equivalently, for every \(\vx\in\mathbb{R}^n\),
\[
\bigl|\,\|(\mW-\eta \operatorname{Ortho}(\mM))\vx\|-\|\mW\vx\|\,\bigr|
\le \eta \|\vx\|.
\]
Orthogonalized updates therefore control the change in activations which in turn is widely thought to improve the stability of optimization for matrix-valued hidden-layer parameters. Muon~\citep{jordan2024muon}, or \textit{MomentUm Orthogonalized by Newton-Schulz}, is a practical optimizer built on this principle. Given a matrix-valued parameter $\mW_t$ and gradient $\mG_t=\nabla_{\mW}L(\mW_t)$, Muon first forms a momentum matrix $\mM_{t+1}$, and then updates $\mW_t$ using an approximate orthogonalization of $\mM_{t+1}$. In its idealized form the Muon update is
\[
\mM_{t+1} = \beta \mM_t + (1-\beta)\mG_t,
\qquad
\mW_{t+1} = \mW_t - \eta_t \text{Ortho}(\mM_{t+1}).
\]
In practice, computing a singular value decomposition at every iteration would be too expensive, so Muon approximates $\operatorname{Ortho}(\mM_{t+1})$ using a small number of Newton--Schulz iterations. Later work has focused on making this approximation stable and efficient at scale, including choices of iteration coefficients, normalization, and mixed-precision behaviour \citep[e.g.][]{amsel2026the,liu2025muonscalablellmtraining} but the core idea of orthogonalizing the momentum matrix remains the same.

\section{Exploring the use of spectral layerwise geometry in SAM}

Rather than proposing a single modification to SAM, in this work we study a variety of SAM optimizers obtained by varying the geometry used in the inner perturbation step, the outer update step, or both. In particular, motivated by the recent empirical success of orthogonalized optimization methods such as Muon, we investigate the effect of introducing \textit{layerwise spectral geometry} into the SAM pipeline. Concretely, we compare inner perturbations generated either by the Euclidean norm on a flattened vector space, as in standard SAM, versus a layerwise spectral norm which we describe in Section~\ref{subsec:layerwise}. Concurrently, we also vary the outer step optimizer between AdamW, an elementwise optimizer commonly used in SAM's outer step~\citep{samforvit}, and Muon. While the use of Muon in the outer step is a natural progression given that SAM is often paired successfully with the same broad classes of stochastic optimizers used for standard empirical risk minimization, it is not obvious that Muon should necessarily enjoy the same success when it comes to minimizing SAM's robust loss surrogate. Finally, we emphasize that our aim is not to advocate a single algorithm, but rather to gain insight through experiments as to how these different geometric choices in the inner and outer step of SAM interact, and provide a practical reference for the effect of layerwise or matrix-aware optimization in SAM. We hope this will motivate and support further research in this direction.

\subsection{Layerwise geometry}\label{subsec:layerwise}
Instead of thinking of our parameters as a flattened vector \(\Theta \cong \R^p\), we explicitly decompose the parameter space into layerwise blocks:
\[
\Theta
=
\R^{m_1\times m_2}\times \R^{m_2\times m_3}\times \cdots \times \R^{m_{L-1}\times m_L}\times \R^q.
\]
We write $\vtheta = (\mW^{(1)},\dots,\mW^{(L-1)},\vv) \in \Theta$ where each \(\mW^{(\ell)}\in\R^{m_\ell\times m_{\ell+1}}\) is a matrix-valued parameter and \(\vv\in\R^q\) collects all remaining non-matrix parameters, such as biases and normalization parameters. As is standard practice in Muon, convolutions are reshaped as local-patch-operator matrices \citep{jordan2024muon}. Given norms \(\|\cdot\|_{(\ell)}\) on the matrix-valued layers and a norm \(\|\cdot\|_{(L)}\) on \(\R^q\), we define, for perturbations
\[
\vveps=(\mE^{(1)},\dots,\mE^{(L-1)},\ve)\in\Theta,
\]
the product norm \(\|\cdot\|_{\times,r}\) as the \(\ell_r\)-norm, \(1\le r\le \infty\), over the norms of all blocks:
\begin{equation}\label{eq:block-aggregation-norm}
\|\vveps\|_{\times,r}
=
\left\|
\bigl(
\|\mE^{(1)}\|_{(1)},\dots,\|\mE^{(L-1)}\|_{(L-1)},\|\ve\|_{(L)}
\bigr)
\right\|_r.
\end{equation}
Equivalently, for \(1\le r<\infty\),
\[
\|\vveps\|_{\times,r}
=
\left(
\sum_{\ell=1}^{L-1}
\|\mE^{(\ell)}\|_{(\ell)}^r
+
\|\ve\|_{(L)}^r
\right)^{1/r},
\]
while
\[
\|\vveps\|_{\times,\infty}
=
\max\left\{
\|\mE^{(1)}\|_{(1)},\dots,
\|\mE^{(L-1)}\|_{(L-1)},
\|\ve\|_{(L)}
\right\}.
\]
The block norms \(\|\cdot \|_{(\ell)}\) determine how a perturbation's magnitude is measured within each layer, while the outer \(\ell_r\)-aggregation determines how perturbation budgets are distributed across layers.

For notational ease we write $\mG^{(\ell)}(\vtheta)
:=\nabla_{\mW^{(\ell)}}L(\vtheta)$ and $\vg(\vtheta) :=\nabla_{\vv}L(\vtheta)$ for the loss gradients of the blocks. We identify \(DL(\vtheta)\in\Theta^*\) with the block gradients via the canonical Frobenius-Euclidean pairing: letting $\langle \mA, \mB \rangle_F = \text{Tr}(\mA^T \mB)$ denote the Frobenius inner product, then
\[
DL(\vtheta)[\vveps]
=
\sum_{\ell=1}^{L-1}
\langle \mG^{(\ell)}(\vtheta),\mE^{(\ell)}\rangle_F
+
\langle \vg(\vtheta),\ve\rangle.
\]
Therefore, if \(r^*\) is the Hölder conjugate of \(r\)\footnote{Recall \(r^*\) is the Hölder conjugate of \(r\) iff $\tfrac{1}{r} + \tfrac{1}{r^*} = 1$}, with the conventions \(1^*=\infty\) and \(\infty^*=1\), the dual product norm evaluated at the differential is
\begin{equation} \label{eq:dual-prod}
\|DL(\vtheta)\|_{\times,r,*}
=
\left\|
\bigl(
\|\mG^{(1)}(\vtheta)\|_{(1),*},
\dots,
\|\mG^{(L-1)}(\vtheta)\|_{(L-1),*},
\|\vg(\vtheta)\|_{(L),*}
\bigr)
\right\|_{r^*}.
\end{equation}
Observe the flattened Euclidean norm used in classic SAM is recovered from~\eqref{eq:block-aggregation-norm} by using Frobenius norms $\|\cdot\|_{(\ell)}=\|\cdot\|_F$ for $\ell \in [L-1]$, the Euclidean norm, \(\|\cdot\|_{(L)}=\|\cdot\|_2\) on the remaining flattened parameters, and the Euclidean aggregation norm \(r=2\). Motivated by Muon, we pay particular attention to the \(\ell_\infty\) aggregation geometry.
When \(r=\infty\) we have \(r^*=1\), and therefore Eq.~\eqref{eq:dual-prod} becomes
\[
\|DL(\vtheta)\|_{\times,\infty,*}
=
\sum_{\ell=1}^{L-1}
\|\mG^{(\ell)}(\vtheta)\|_{(\ell),*}
+
\|\vg(\vtheta)\|_{(L),*}.
\]
Let \(\partial \|\cdot\|_{\times,\infty,*}(DL(\vtheta))\) denote the subdifferential of the dual norm as a function on \(\Theta^*\), evaluated at the covector \(DL(\vtheta)\). Using the finite-dimensional identification \((\Theta^*)^*\cong \Theta\), its elements are identified with perturbations in parameter space. Since the dual product norm is a sum of convex functions, each acting on a disjoint parameter block, the standard subdifferential rule for sums gives
\[
\partial \|\cdot\|_{\times,\infty,*}\bigl(DL(\vtheta)\bigr)
=
\partial\|\cdot\|_{(1),*}\bigl(\mG^{(1)}(\vtheta)\bigr)
\times \cdots \times
\partial\|\cdot\|_{(L-1),*}\bigl(\mG^{(L-1)}(\vtheta)\bigr)
\times
\partial\|\cdot\|_{(L),*}\bigl(\vg(\vtheta)\bigr).
\]
Thus, under the \(r=\infty\) block-aggregation geometry, the steepest ascent perturbation may be computed independently for each layer by choosing a member of the subdifferential of the corresponding blockwise dual norm.
Motivated by Muon, we specialize to the spectral norm, $\|\cdot\|_{(\ell)}=\|\cdot\|_{2\to 2}$ for all $\ell \in [L-1]$, and the Euclidean norm \(\|\cdot\|_{(L)}=\|\cdot\|_2\) on the remaining parameters. Recall the dual norm of the spectral norm is the nuclear norm:
the nuclear norm is differentiable at a matrix argument if and only if said matrix is full rank. Therefore, in general, the steepest direction need not be unique. As highlighted in Subsection~\ref{subsec:muon}, a canonical choice of nuclear-norm subgradient is the polar factor \(\mathrm{Ortho}(\mG^{(\ell)}(\vtheta))\).
Therefore, for \(\vg(\vtheta)\neq 0\), the idealized layerwise spectral unit perturbation is
\begin{equation}\label{eq:spectral-layer-perturbation}
\vveps^*(\vtheta)
=
\left(
\mathrm{Ortho}\!\bigl(\mG^{(1)}(\vtheta)\bigr),
\dots,
\mathrm{Ortho}\!\bigl(\mG^{(L-1)}(\vtheta)\bigr),
\frac{\vg(\vtheta)}{\|\vg(\vtheta)\|_2 }
\right).
\end{equation}
The spectral norm used for matrix blocks and the Euclidean norm used for non-matrix parameters measure perturbation size in different ''units''. In principle, this suggests using separate perturbation radii, or equivalently separate radius scaling factors, for matrix and non-matrix parameter groups. 

This leads to the practical spectral perturbation used in Section~\ref{sec:experiments}:
\begin{equation}\label{eq:practical-spectral-layer-perturbation}
\widetilde{\vveps^*}(\vtheta)
=
\left(
\mathrm{NS}_k\!\bigl(\mG^{(1)}(\vtheta)\bigr),
\dots,
\mathrm{NS}_k\!\bigl(\mG^{(L-1)}(\vtheta)\bigr),
\alpha_g\frac{\vg(\vtheta)}{\|\vg(\vtheta)\|_2}
\right),
\end{equation}
where $\mathrm{Ortho}(\cdot)$ is approximated by a $k$-step Newton--Schulz
iteration $\mathrm{NS}_k$ \citep{jordan2024muon}, and $\alpha_g\geq 0$ sets the
perturbation budget of the Euclidean-measured non-matrix block relative to the
spectral-measured matrix blocks.

\subsection{Experimental variants}
We now instantiate Algorithm~\ref{alg:general-SAM} with concrete choices for the inner perturbation geometry and the outer update rule, with the specific goal of exploring how matrix-aware geometry \citep[in the sense of ][]{bernstein2025modularduality} may be incorporated into SAM. The variants we explore are summarized in Table~\ref{tab:method-grid}, including their shortened names used for brevity throughout the paper.

For the inner step, we consider two main choices. The first is standard Euclidean
SAM under the global $\ell_2$ geometry on the full parameter vector, with inner
perturbation $\rho\vveps_t = \rho\,\mathbf g_t/\|\mathbf g_t\|_2$ for the loss
gradient $\mathbf g_t$ over all parameters. We refer to this variant by the prefix
``\textsc{SAM-}''. The second is a layerwise variant that combines the $r=\infty$
block aggregation with the spectral norm on each matrix-valued parameter block,
giving the perturbation of Eq.~\eqref{eq:practical-spectral-layer-perturbation}.
We refer to it by the prefix ``\textsc{SpecSAM-}''. The spectral geometry applies
only to matrix-valued blocks, with all remaining parameters perturbed under the
Euclidean geometry. Note that this set of matrices is larger than the Muon group
used in the outer step; see Appendix~\ref{sec:app:routing} for details.
As an ablation, we additionally consider a ``\textsc{LayerSAM-}'' control that uses
the same $r=\infty$ layerwise aggregation as \textsc{SpecSAM-} but replaces the
spectral block norm with the Frobenius norm. This isolates the effect of per-layer
budgeting from that of the choice of block norm. We study it in
Appendix~\ref{sec:app:layersam}.

For the outer step, we consider AdamW, SGDW, and Muon. AdamW is used for the
vision transformer (ViT) and SGDW for the convolutional network (CNN); throughout,
we write \textsc{AdamW/SGDW} for this architecture-dependent choice. In the
\textsc{AdamW/SGDW} variants, the direction map $\mathcal{D}_t$ of
Algorithm~\ref{alg:general-SAM} is AdamW/SGDW applied to all parameters. In the
\textsc{Muon} variants, we follow the usual hybrid routing, in which Muon updates
only the matrix-valued hidden-layer parameters,\footnote{Some matrices are excluded
from the Muon group and updated with AdamW; see Appendix~\ref{sec:app:routing}.}
while the remaining parameters are updated by AdamW/SGDW. With the prefixes
introduced above, this yields four methods: \textsc{SAM-AdamW/SGDW},
\textsc{SAM-Muon}, \textsc{SpecSAM-AdamW/SGDW}, and \textsc{SpecSAM-Muon}. Together
with the two unperturbed baselines, these form the grid of
Table~\ref{tab:method-grid}.

\begin{table}[t]
\centering
\small
\setlength{\tabcolsep}{7pt}
\renewcommand{\arraystretch}{1.55}
\begin{tabular}{
|>{\centering\arraybackslash}p{0.22\linewidth}
|>{\centering\arraybackslash}p{0.32\linewidth}
|>{\centering\arraybackslash}p{0.32\linewidth}|
}
\hline
\multirow{2}{*}{\textbf{Inner Step}} & \multicolumn{2}{c|}{\textbf{Outer Step}} \\
\cline{2-3}
& \textbf{AdamW/SGDW} & \textbf{Muon} \\
\hline
\textbf{None}
&
$\begin{gathered}
\textsc{AdamW/SGDW}\\
\vtheta_{t+1} = \mathrm{AdamW/SGDW}(\vtheta_t,\mathbf g_t)
\end{gathered}$
&
$\begin{gathered}
\textsc{Muon}\\
\vtheta_{t+1} = \mathrm{Muon}(\vtheta_t,\mathbf g_t)
\end{gathered}$ \\
\hline
\textbf{Euclidean}
&
$\begin{gathered}
\textsc{SAM-AdamW/SGDW}\\
\vveps_t = \mathbf g_t/\|\mathbf g_t\|_2 \\
\vtheta_{t+1} = \mathrm{AdamW/SGDW}(\vtheta_t,\phi_{\rho,t})
\end{gathered}$
&
$\begin{gathered}
\textsc{SAM-Muon}\\
\vveps_t = \mathbf g_t/\|\mathbf g_t\|_2 \\
\vtheta_{t+1} = \mathrm{Muon}(\vtheta_t,\phi_{\rho,t})
\end{gathered}$ \\
\hline
\textbf{Spectral}
&
$\begin{gathered}
\textsc{SpecSAM-AdamW/SGDW}\\
\vveps_t = \widetilde{\vveps^*}(\vtheta_t)\ \text{from}\ \eqref{eq:practical-spectral-layer-perturbation} \\
\vtheta_{t+1} = \mathrm{AdamW/SGDW}(\vtheta_t,\phi_{\rho,t})
\end{gathered}$
&
$\begin{gathered}
\textsc{SpecSAM-Muon}\\
\vveps_t = \widetilde{\vveps^*}(\vtheta_t)\ \text{from}\ \eqref{eq:practical-spectral-layer-perturbation} \\
\vtheta_{t+1} = \mathrm{Muon}(\vtheta_t,\phi_{\rho,t})
\end{gathered}$ \\
\hline
\end{tabular}
\caption{
    A summary of the experimental design space. Here $\mathbf g_t=\nabla L(\vtheta_t)$ is the gradient and $\phi_{\rho,t}=\nabla L(\vtheta_t+\rho\vveps_t)$ the robust-loss gradient at step $t$. \textsc{AdamW}, \textsc{SGDW}, and \textsc{Muon} denote the procedure that converts a (perturbed) gradient signal into a parameter update with the corresponding outer optimizer, i.e. the direction map $\mathcal{D}_t$ of Algorithm~\ref{alg:general-SAM} followed by the parameter step.}
\label{tab:method-grid}
\end{table}

\section{Experiments}\label{sec:experiments}
To evaluate the effect of matrix-aware geometry in the inner and outer steps, we evaluate the proposed SAM variants on two common SAM settings: ImageNet-1K \citep{imagenet15russakovsky} classification with a vision transformer and a convolutional neural network. Across both settings, we compare the SAM variants against their non-SAM baselines, AdamW/SGDW and Muon. Full experimental details are given in Appendix~\ref{sec:app:experimentdetails}. In addition to ImageNet-1K Top-1 validation accuracy, we report the ''ReaL'' accuracy for ImageNet \citep{beyer2020imagenet}, which replaces the original labels with multi-label sets for each image, marking the model as correct if it predicts any label in the set (a framework designed to address label ambiguity). We also assess the Top-1 accuracy on ImageNet-R \citep{hendrycks2021facesrobustnesscriticalanalysis}, an out-of-distribution 30k image test set of artistic renditions of ImageNet images. We additionally perform diagnostic experiments on a small ViT trained on CIFAR-100 \citep{krizhevsky2009learning}, sweeping the SAM radius and analysing sharpness, effective rank, and generalization gap. The final vision transformer experiments as presented, including tuning and all algorithms, took roughly 4000 GPU-hours to run, while the ResNet experiments took 2000 GPU-hours. Models are trained on 8 H100 GPUs and we use the m-sharpness perturbation for SpecSAM and SAM, i.e. the parameter is perturbed independently on each GPU, and only the outer step gradient is accumulated~\citep{foret2020SAM}. The entire project, including preliminary and diagnostic experiments, used about 10,000 hours of H100 96GB GPU compute. Experiments were conducted using the Isambard-AI National AI
Research Resource~\cite{mcintoshsmith2024isambard}.

\subsection{ImageNet-1K results}\label{sec:exp:imagenet}
We train ViT-Small/16 (see Table~\ref{tab:arch-vit} for architectural details) for $300$ epochs following \citet{beyer2022betterplainvitbaselines, samforvit}, and ResNet-50 (Table~\ref{tab:arch-resnet}) for $120$ epochs following the \texttt{timm} \texttt{resnet50} implementation \citep{he2015deepresiduallearningimage,rw2019timm}, using \textsc{AdamW} as the non-Muon outer optimizer for ViT-Small/16 and \textsc{SGDW} for ResNet-50.  To isolate the effects of inner and outer step geometry from unrelated hyperparameter effects, we tune in two stages: first the learning rate and weight decay of the non-SAM AdamW/SGDW and Muon baselines, then the perturbation radius parameters for each SAM variant with the learning rate and weight decay fixed to that of its outer optimizer. Full configurations are given in Appendix~\ref{sec:app:hpsearch}. Throughout the ImageNet-1K experiments, the number of Newton-Schulz iterations for the Muon outer step is $5$, while the spectral inner step uses $3$ iterations.

The first half of Table~\ref{tab:main} contains the results for ViT-Small/16. Without an inner perturbation, Muon improves only modestly over AdamW ($74.26\to 74.95$). A Euclidean ($\ell_2$) inner step then contributes almost the same gain regardless of the outer optimizer (roughly $+4.9$ for AdamW and $+5.0$ for Muon):
 here the inner and outer effects are essentially \emph{additive}. The spectral inner step behaves differently: it amplifies the advantage of Muon. \textsc{SpecSAM-AdamW} is in fact the weakest SAM method ($78.71$, below even \textsc{SAM-AdamW}), whereas \textsc{SpecSAM-Muon} attains the best validation accuracy of all methods ($80.23$). The spectral perturbation improves validation accuracy by 5.28 percentage points with Muon, compared with 4.45 points with AdamW. This suggests that the spectral inner geometry is more effective when paired with a similarly motivated outer step. The out-of-distribution ImageNet-R results point in the same direction
($42.18$ for \textsc{SpecSAM-Muon}), though at three seeds the top three
methods there are not significantly different under the reported test. Interestingly, both spectral inner step variants reach \emph{lower} train accuracy than their Euclidean counterparts, perhaps implying that the spectral inner geometry more harshly penalizes memorizing the training data.

\begin{table}[t]
\centering
\small
\setlength{\tabcolsep}{6pt}
\begin{tabular}{cccrrrr}
\toprule
Name & Inner Step & Outer Step & Val Top-1 & Train Top-1 & ReaL & ImageNet-R\\
\midrule
\multicolumn{7}{c}{\emph{ViT-Small/16, 300 epochs}}\\
\midrule
\textsc{SpecSAM-Muon}  & Spectral  & Muon  & $\mathbf{80.23}_{\pm 0.06}$ & $89.84_{\pm 0.05}$ & $\mathbf{86.26}_{\pm 0.08}$ & $\mathbf{42.18}_{\pm 0.24}$\\
\textsc{SAM-Muon}      & Euclidean & Muon  & $79.97_{\pm 0.06}$          & $90.68_{\pm 0.08}$ & $86.00_{\pm 0.03}$          & $\mathbf{41.68}_{\pm 0.56}$\\
\textsc{SAM-AdamW}     & Euclidean & AdamW & $79.17_{\pm 0.35}$          & $91.02_{\pm 0.52}$ & $85.12_{\pm 0.23}$          & $\mathbf{39.17}_{\pm 1.41}$\\
\textsc{SpecSAM-AdamW} & Spectral  & AdamW & $78.71_{\pm 0.08}$          & $89.34_{\pm 0.23}$ & $84.95_{\pm 0.15}$          & $38.13_{\pm 0.53}$\\
\textsc{Muon}          & None      & Muon  & $74.95_{\pm 0.16}$          & $93.38_{\pm 1.93}$ & $80.79_{\pm 0.11}$          & $34.35_{\pm 0.35}$\\
\textsc{AdamW}         & None      & AdamW & $74.26_{\pm 0.09}$          & $92.41_{\pm 0.61}$ & $80.09_{\pm 0.06}$          & $32.68_{\pm 0.19}$\\
\midrule
\multicolumn{7}{c}{\emph{ResNet-50, 120 epochs}}\\
\midrule
\textsc{SpecSAM-Muon} & Spectral  & Muon & $\mathbf{78.55}_{\pm 0.06}$ & $93.67_{\pm 0.01}$ & $\mathbf{84.76}_{\pm 0.10}$ & $\mathbf{38.12}_{\pm 0.45}$\\
\textsc{SAM-SGDW}     & Euclidean & SGDW & $77.99_{\pm 0.14}$          & $92.69_{\pm 0.08}$ & $84.26_{\pm 0.07}$          & $\mathbf{37.50}_{\pm 0.34}$\\
\textsc{SAM-Muon}     & Euclidean & Muon & $77.84_{\pm 0.03}$          & $95.24_{\pm 0.15}$ & $83.90_{\pm 0.05}$          & $\mathbf{37.92}_{\pm 0.23}$\\
\textsc{SpecSAM-SGDW} & Spectral  & SGDW & $77.77_{\pm 0.15}$          & $93.16_{\pm 0.02}$ & $84.05_{\pm 0.06}$          & $36.75_{\pm 0.13}$\\
\textsc{SGDW}         & None      & SGDW & $77.35_{\pm 0.07}$          & $93.33_{\pm 0.11}$ & $83.55_{\pm 0.13}$          & $36.04_{\pm 0.12}$\\
\textsc{Muon}         & None      & Muon & $77.05_{\pm 0.07}$          & $96.35_{\pm 0.01}$ & $83.17_{\pm 0.12}$          & $34.92_{\pm 0.24}$\\
\bottomrule
\end{tabular}
\caption{ImageNet-1K results, reported as mean$_{\pm\text{std}}$ over three seeds and
ordered within each panel by Val Top-1. In each column except Train Top-1, we bold the method with the
highest mean together with methods for which a two-sided Welch
$t$-test does not reject equality with the highest-mean method at
the $5\%$ level. Bold entries therefore indicate that no statistically
significant difference from the highest-mean method was detected.}
\label{tab:main}
\end{table}

\begin{figure}[t]
\centering
\includegraphics[width=0.8\textwidth]{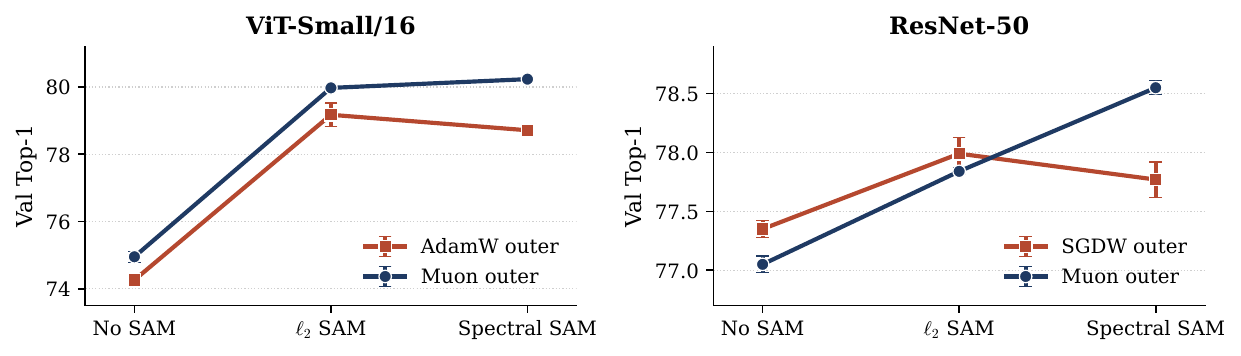}
\caption{Inner step geometry vs.\ outer optimizer on ImageNet-1K (left:
ViT-Small/16, right: ResNet-50).}
\label{fig:inner-outer}
\end{figure}

The second half of Table~\ref{tab:main} shows analogous results for ResNet-50. Without a SAM inner step, Muon slightly trails the standard SGDW baseline ($77.05$ vs $77.35$). A Euclidean inner step again adds a roughly constant gain across outer optimizers ($+0.64$ for SGDW, $+0.79$ for Muon), leaving \textsc{SAM-SGDW} ($77.99$) marginally ahead of \textsc{SAM-Muon} ($77.84$). The spectral inner step once more amplifies Muon's advantage: it lifts Muon by $1.50$ to $78.55$, the best result overall, but SGDW by only $0.42$ to $77.77$, flipping the ordering so that the Muon pairing now clearly wins. \textsc{SpecSAM-Muon} also attains the highest ImageNet-ReaL accuracy, and
leads on ImageNet-R as well, although as on ViT-Small/16 the ImageNet-R
margins are within seed noise. 
Figure~\ref{fig:inner-outer} summarizes this interaction across both
architectures: the $\ell_2$ inner step shifts both outer optimizers by a
near-constant amount, whereas the spectral inner step selectively benefits Muon.


\subsection{CIFAR-100 diagnostics}\label{sec:exp:rho-sweep}
 
\begin{figure}[t]
\centering
\includegraphics[width=1.0\linewidth]{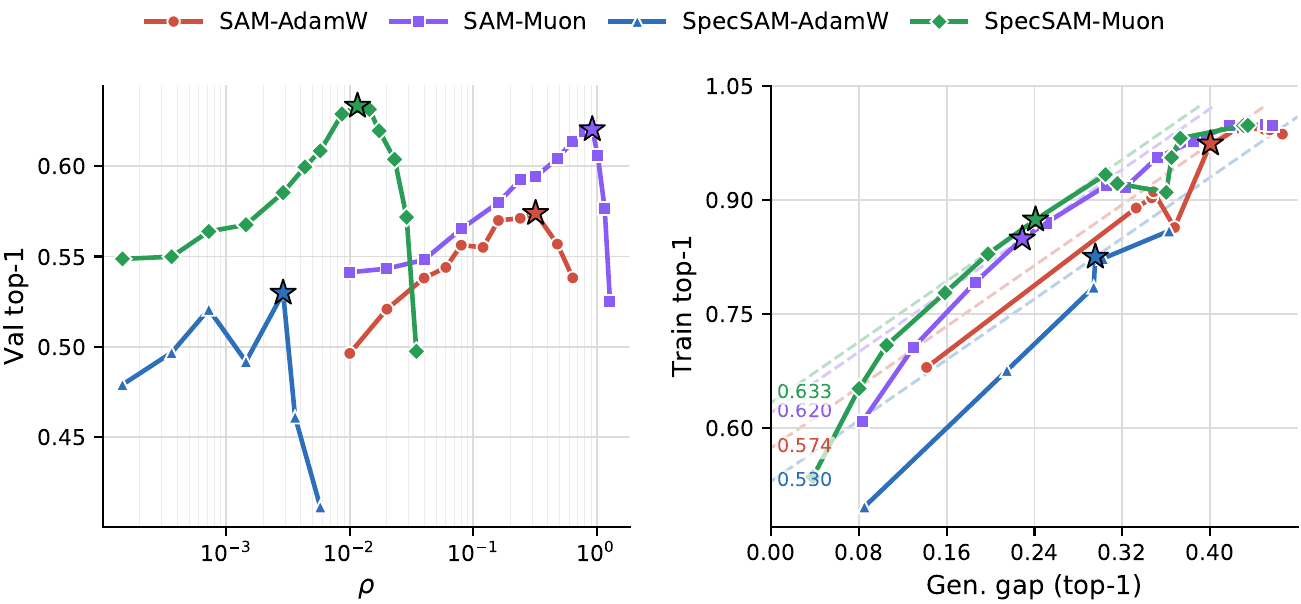}
\caption{Sensitivity to SAM training radius $\rho$ on CIFAR-100 with ViT-Tiny/4. \textbf{Left:} best validation top-1 vs.\ $\rho$. \textbf{Right:} train top-1 vs.\ generalization gap, with $\rho$ as an implicit variable along each curve; dashed diagonals are constant val-top-1 contours since $\mathrm{val}=\mathrm{train}-\mathrm{gap}$. Stars mark the best-validation checkpoint per method.}
\label{fig:rho-sweep}
\end{figure}

The ImageNet tables compare a single tuned perturbation radius per method. To see the interaction across $\rho$, we sweep $\rho$ for the four SAM variants on CIFAR-100 with ViT-Tiny/4 \citep{beyer2022betterplainvitbaselines} trained for $120$ epochs, with all four methods reusing one learning rate and weight decay selected by an AdamW grid search. Figure~\ref{fig:rho-sweep} summarizes validation top-1 sensitivity to $\rho$, and the train-accuracy--generalization-gap relationship.
 
The left panel of Figure~\ref{fig:rho-sweep} shows that the Euclidean inner steps prefer a larger $\rho$ than the spectral inner steps.
The right panel plots the relationship between train accuracy and the generalization gap (train minus validation top-1 accuracy) from the same set of results. Because validation top-1 equals train top-1 minus the gap, the dashed diagonals are constant-validation contours, and a method is ''better'' when it reaches a higher contour at its optimal $\rho$. This contour position reflects how efficiently a method converts loss minimization into a reduced generalization gap, which depends on two factors: how well the outer optimizer minimizes the robust loss, and how much regularization the inner perturbation delivers per unit of training accuracy given up. \textsc{SpecSAM-AdamW} fares worst, sacrificing too much training accuracy for too little regularization in return. \textsc{SAM-Muon} and \textsc{SpecSAM-Muon} both reach the highest contour, retaining substantial train accuracy while keeping the gap moderate, with \textsc{SpecSAM-Muon} handling the trade-off slightly better.

Effective-rank diagnostics on the same sweep, reported in full in Appendix~\ref{sec:app:rank-diagnostics}, give further insights into how spectral geometry interacts with SAM: in particular, Muon outer updates maintain higher weight, gradient, and feature ranks across the radius range.
Inner perturbations at large $\rho$, by contrast, reduce all three rank quantities; this phenomenon was previously discussed by~\citet{Andriushchenko2023SAMlowrank}, who posit that SAM may lead to low rank features.

\section{Related Work}

Sharpness-Aware Minimization (SAM) was introduced by \citet{foret2020SAM} as a two-step procedure for improving generalization by optimizing a local worst-case loss. Subsequent work has studied why SAM improves generalization \citep[e.g.][]{andriushchenko2022understandingsharpnessawareminimization, wen2023doessharpnessawareminimizationminimize} and proposed variants aimed at improving performance, reducing cost, or adapting SAM to particular settings \citep[e.g.][]{lookSAM, zhuang2022surrogategapminimizationimproves, li2024friendlysam,becker2025momentumsamsharpnessawareminimization}.

Most relevant to our work are SAM variants that modify the geometry of the perturbation. ASAM \citep{pmlr-v139-kwon21b} rescales the inner perturbation to reduce sensitivity to parameter rescaling; Fisher SAM \citep{pmlr-v162-kim22f} replaces Euclidean perturbation balls with Fisher-induced ellipsoids; and Riemannian SAM \citep{yun2023riemanniansam} formulates perturbations on Riemannian manifolds. Our work differs in two respects: we study spectral-norm geometry for matrix-valued parameters, and we vary both the inner perturbation geometry and the outer optimizer. Empirically, our strongest results arise when spectral inner perturbations are paired with a matrix-aware outer update. We also distinguish our method from Spectral-SAM \citep{zagitov2025spectralsam}, which targets adversarial robustness by projecting weights onto spectral-norm balls, whereas we modify the geometry of the SAM perturbation and update.

A separate line of work argues that hidden-layer weight matrices should be optimized using matrix-aware geometry rather than flattened Euclidean geometry \citep{bernstein2025modularduality, pethick2025trainingdeeplearningmodels}. This perspective motivates Muon \citep{jordan2024muon}, which applies momentum followed by approximate orthogonalization and can be interpreted as a spectral-norm steepest-descent method for matrix parameters. Muon has recently been adopted in large-scale language-model training \citep{kimiteam2025kimik2openagentic,5team2025glm45agenticreasoningcoding}. To our knowledge, however, this matrix-aware optimization perspective has not previously been studied in the SAM setting.

\section{Conclusion}
\textbf{Summary.} We studied Sharpness-Aware Minimization through a
norm-agnostic template that separates the geometry of the inner
perturbation from that of the outer update, and, motivated by
matrix-aware optimization, introduced a layerwise spectral inner
perturbation for matrix-valued parameters. Across ImageNet-1K
experiments on ViT-Small/16 and ResNet-50, the two geometries interact: a Euclidean inner step shifts both outer optimizers by
a near-constant amount, whereas the spectral inner step selectively
amplifies Muon's advantage. \textsc{SpecSAM-Muon} attains the best validation accuracy on both
architectures ($80.23$ and $78.55$) and the best ImageNet-ReaL accuracy,
while the same perturbation paired with AdamW/SGDW is markedly less
effective. The ordering on out-of-distribution ImageNet-R is the same,
though at three seeds it is not statistically resolvable.
Why the two geometries should interact this way is not settled by
our experiments. One suggestive observation is that Muon sustains
higher weight and feature effective rank across the CIFAR-100
radius sweep, which may be what lets it absorb a spectral
perturbation without losing representational capacity; a proper
account is left to future work.

\textbf{Limitations.} Compute constraints limited us to three seeds for
the main ImageNet results, single runs for some diagnostics, and
non-exhaustive hyperparameter searches. Our evidence is confined to
image classification with two architectures; whether the same inner/outer
interaction holds for language modelling, where Muon is most widely
deployed, remains untested. Finally, like standard SAM, our methods require two
backward passes per step, and the spectral variants add Newton--Schulz
overhead in the inner step; we compare at equal epoch budgets rather
than equal wall-clock time.

\textbf{Future work.} Our results suggest that future SAM variants should consider using Muon for matrix-shaped parameters in the outer step, since it is a simple replacement for AdamW/SGDW and performed strongly in our experiments. Matrix-aware inner geometries also deserve further study, especially when paired with a matrix-aware outer optimizer. Beyond the fixed Euclidean and spectral choices studied here, preliminary experiments provided anecdotal evidence that scheduling or interpolating between inner geometries may be promising. Finally, more theoretical work is needed to understand precisely why certain inner geometries pair better with some optimizers than others.

\section*{Acknowledgments}

The authors acknowledge the use of resources provided by the
Isambard-AI National AI Research Resource (AIRR). Isambard-AI is
operated by the University of Bristol and is funded by the UK
Government’s Department for Science, Innovation and Technology
(DSIT) via UK Research and Innovation; and the Science and
Technology Facilities Council [ST/AIRR/I-A-I/1023].

\bibliography{refs}
\bibliographystyle{plainnat}

\clearpage
\appendix
\section{Experiment details}\label{sec:app:experimentdetails}

\subsection{Architectures}\label{sec:app:architectures}
The main results in Table~\ref{tab:main} are obtained by evaluating the methods listed in Table~\ref{tab:method-grid} on a vision transformer model and a convolutional neural network, both at $224\times 224$ resolution. The transformer model is the ViT-Small/16 from the original ViT family~\citep{dosovitskiy2020vit} with non-overlapping $16\times 16$ patches, modified following \citet{beyer2022betterplainvitbaselines} with fixed two-dimensional sine--cosine positional embeddings, global average pooling over patch tokens rather than a class-token representation, and an MLP classification head. The convolutional baseline is the standard \texttt{timm} \texttt{resnet50} model with no architectural modifications. 
The model used for the CIFAR-100 diagnostics is ViT-Tiny/4, which follows the same overall design as ViT-Small/16 but uses a smaller architectural configuration and is adapted to $32\times 32$ CIFAR images. Tables~\ref{tab:arch-vit} and~\ref{tab:arch-resnet} list the main numerical specifications for reference.

\begin{table}[htbp]
\centering
\begin{tabular}{lcc}
\hline
Specification & ViT-Small/16 & ViT-Tiny/4 \\
\hline
Input resolution & $224\times224$ & $32\times32$ \\
Patch size & $16\times16$ & $4\times4$ \\
Patch stride & 16 & 4 \\
Number of patch tokens & 196 & 64 \\
Embedding width & 384 & 192 \\
Depth & 12 & 12 \\
Attention heads & 6 & 3 \\
Head dimension & 64 & 64 \\
MLP ratio & 4 & 4 \\
Hidden expansion & 1536 & 768 \\
Classifier output dim & 1000 & 100 \\
\hline
\end{tabular}
\caption{Architecture specifications of ViT-Small/16 and ViT-Tiny/4.}
\label{tab:arch-vit}
\end{table}

\begin{table}[htbp]
\centering
\begin{tabular}{lc}
\hline
Specification & ResNet-50 \\
\hline
Input resolution &  $224\times224$ \\
Stem kernel size & 7 \\
Stem stride & 2 \\
Stem width & 64 \\
Total layers & 50 \\
Stage/block layout & 3/4/6/3 \\
Stage widths & 256/512/1024/2048 \\
Classifier output dim & 1000 \\
\hline
\end{tabular}
\caption{Architecture specifications of ResNet-50.}
\label{tab:arch-resnet}
\end{table}

\subsection{Datasets and preprocessing}\label{sec:app:datasets}
The main experiments use ImageNet-1K~\citep{imagenet15russakovsky}, with the official training split for optimization, the validation split for evaluation, and inception-style augmentation~\citep{Szegedy2016inception}. As in other SAM works~\citep{samforvit,zhuang2022surrogategapminimizationimproves,du2022efficient}, we deliberately do not use Mixup, CutMix, RandAugment, so that the comparison reflects the optimization effect rather than the augmentation recipe. ImageNet-ReaL accuracy~\citep{beyer2020imagenet} is computed on the same validation images using the multi-label ReaL annotations: a prediction is correct if it matches any label in the ReaL set. ImageNet-R~\citep{hendrycks2021facesrobustnesscriticalanalysis} is an out-of-distribution benchmark of approximately $30\mathrm{k}$ rendition images covering $200$ ImageNet classes, evaluated with the same preprocessing as the validation split. The diagnostic experiments use CIFAR-100~\citep{krizhevsky2009learning} with the standard $50\mathrm{k}/10\mathrm{k}$ split, random cropping with $4$-pixel padding and horizontal flipping during training.

\subsection{Training protocol}\label{sec:app:training}
ViT-Small/16 follows the 300-epoch recipe of \citet{beyer2022betterplainvitbaselines,samforvit}, using a cosine learning-rate schedule with linear warmup. ResNet-50 is trained for 120 epochs with the same family of schedules, matching the standard \texttt{timm} \texttt{resnet50} reference setup. CIFAR-100 ViT-Tiny/4 runs use the same family of cosine-with-warmup schedules over 120 epochs. For ResNet-50 and ViT-Small/16, we use a batch size of \(4096\) and 6000 warmup steps for ResNet and 10000 warmup steps for ViT, while for ViT-Tiny/4 we use a batch size of \(512\) and 1000 warmup steps.
Following \citet{samforvit}, ViT experiments use binary cross-entropy loss with label smoothing, while ResNet-50 uses standard softmax cross-entropy.

For all experiments, the model dropout rate is 0.1 and the loss label smoothing is 0.1. We do not apply drop path to the ViT models. AdamW uses \((\beta_1,\beta_2)=(0.9,0.999)\), SGDW uses momentum \(0.9\), and Muon uses momentum \(0.95\). ViT-Small/16 and ResNet-50 are trained on 8 H100 GPUs, while ViT-Tiny/4 is trained on 1 H100 GPU. All reported metrics are evaluated at the best-validation checkpoint; training top-1 accuracy is evaluated using the same transform as the validation split, without training-time augmentation. The results reported in Table~\ref{tab:main} are based on three random seeds.

\subsection{Optimizer routing and implementation}\label{sec:app:routing}
The methods of Table~\ref{tab:method-grid} fix the inner perturbation geometry and the outer optimizer family but leave parameter routing implicit; we make these choices explicit here. For \textsc{AdamW/SGDW} and the \textsc{*-AdamW/SGDW} variants, all parameters are updated by AdamW / SGDW respectively. For \textsc{Muon} and the \textsc{*-Muon} variants, parameters are split into a Muon group containing the hidden-layer matrix-valued backbone parameters and an AdamW/SGDW group containing the rest. On ViT-Small/16, the classification head and the patch-embedding layer are excluded from the Muon group; on ResNet-50, convolutional kernels are reshaped to two-dimensional matrices and routed to Muon with the final classification head and the first convolutional layer again excluded. We use a Moonlight-style update scaling for Muon~\citep{liu2025muonscalablellmtraining} to match update RMS of AdamW, and Newton--Schulz iterations for the outer Muon orthogonalization are fixed at $5$. The \textsc{SpecSAM} inner step uses a strictly broader perturbation set than the outer Muon group: all matrix-valued parameters are perturbed, including those routed to AdamW/SGDW for the outer step (note that this includes embeddings, the classification layer, and the low-dimension convolutions), while non-matrix parameters use the Euclidean \textsc{SAM} variant with a separately tuned radius (see Appendix~\ref{sec:app:hpsearch}). The orthogonalization in the inner step is approximated by Newton--Schulz iterations, with 3 steps used in the main reported numbers for ImageNet experiments, and five steps for CIFAR-100 experiments. The spectral perturbation radii \(\rho\) for the output layer of ViT-Tiny/4 are \(\sqrt{1.92}\) times as large as those for the other parameters.

\subsection{Hyperparameter search}\label{sec:app:hpsearch}

We tune hyperparameters separately for each architecture using a two-stage grid search. To reduce the computational cost, all hyperparameter searches use half of the training budget of the final experiments, namely 150 epochs for ViT-Small/16 and 60 epochs for ResNet-50. We first tune the learning rate parameters and weight decay for the non-SAM baselines. For each sharpness-aware variant, we retain the learning rate and weight decay parameters selected for the corresponding outer optimizer and tune the perturbation radius parameters. This protocol prevents a SAM variant from benefiting solely from a learning-rate choice unavailable to its corresponding non-SAM baseline.

\paragraph{Notation.}
 Throughout this section, $\mathrm{LR}$
and $\mathrm{WD}$ denote the base learning rate and the weight decay, respectively, and $\rho$
 denotes the perturbation radius of the sharpness-aware update. As described in Appendix~\ref{sec:app:routing}, when Muon is the outer optimizer the parameters are split into a matrix branch, updated by Muon, and a non-matrix branch, updated by AdamW/SGDW; these two branches use separate learning rates, which we denote by $\mathrm{mLR}$ and $\mathrm{nLR}$ respectively. Similarly, the spectral perturbation uses separate radii for the matrix and non-matrix parameter groups, which we denote by $\rho_m$
 and $\rho_n$, i.e., if \(\rho = \rho_m\) in Table~\ref{tab:method-grid}, \(\rho_n=\alpha_g\rho_m\) according to \eqref{eq:practical-spectral-layer-perturbation}. 

\begin{table}[t]
\centering
\small
\setlength{\tabcolsep}{6pt}
\renewcommand{\arraystretch}{1.2}
\begin{tabular}{@{}lll@{}}
\toprule
Method & Hyperparameter & Search grid \\
\midrule
AdamW
& $\mathrm{LR}$
& $10^{-3}\{1,\sqrt{3},3,3\sqrt{3},9\}$ \\
& $\mathrm{WD}$
& $10^{-1}\{\tfrac{1}{3},\,1,\,3\}$ \\
\addlinespace
Muon
& $\mathrm{mLR},\mathrm{nLR}$
& $10^{-3}\{1,\sqrt{3},3,3\sqrt{3},9\}$ \\
& $\mathrm{WD}$
& $10^{-1}\{\tfrac{1}{3},\,1,\,3\}$ \\
\addlinespace
SAM-AdamW
& $\rho$
& $10^{-1}\{3,\sqrt{21},7,\tfrac{7\sqrt{21}}{3}\}$ \\
\addlinespace
SAM-Muon
& $\rho$
& $10^{-1}\{3,\sqrt{21},7,\tfrac{7\sqrt{21}}{3}\}$ \\
\addlinespace
SpecSAM-AdamW
& $\rho_m$
& $5.1\times10^{-4}\{\tfrac{9\sqrt{21}}{49},\tfrac{9}{7},\tfrac{3\sqrt{21}}{7},3\}$ \\
& $\rho_n$
& $10^{-3}\{0,0.1,0.1\sqrt{10},1,\sqrt{10},10\}$ \\
\addlinespace
SpecSAM-Muon
& $\rho_m$
& $5.1\times10^{-4}\{\tfrac{9\sqrt{21}}{49},\tfrac{9}{7},\tfrac{3\sqrt{21}}{7},3\}$ \\
& $\rho_n$
& $10^{-3}\{0,0.1,0.1\sqrt{10},1,\sqrt{10},10\}$ \\
\bottomrule
\end{tabular}
\caption{Hyperparameter search grids for ViT-Small/16. The sharpness-aware
variants inherit the learning rate and weight decay selected for their base
optimizer, so only the perturbation-radius grids are listed for them.}
\label{tab:vit-search-grids}
\end{table}

\begin{table}[t]
\centering
\small
\setlength{\tabcolsep}{6pt}
\renewcommand{\arraystretch}{1.2}
\begin{tabular}{@{}lll@{}}
\toprule
Method & Hyperparameter & Search grid \\
\midrule
SGDW
& $\mathrm{LR}$
& $1.6\{\tfrac{1}{3},\tfrac{1}{\sqrt{3}},1,\sqrt{3},3\}$ \\
& $\mathrm{WD}$
& $10^{-3}\{\tfrac{1}{3},1,3\}$ \\
\addlinespace
Muon
& $\mathrm{mLR},\mathrm{nLR}$
& $10^{-2}\{0.2,0.5,1,2,5\}$ \\
& $\mathrm{WD}$
& $10^{-1}\{\tfrac{1}{3},1,3\}$ \\
\addlinespace
SAM-SGDW
& $\rho$
& $10^{-1}\{0.125,0.25,0.5,1,2\}$ \\
\addlinespace
SAM-Muon
& $\rho$
& $10^{-1}\{0.5,1,2,4,8\}$ \\
\addlinespace
SpecSAM-SGDW
& $\rho_m$
& $10^{-3}\{0.2,0.5,1,2,5\}$ \\
& $\rho_n$
& $10^{-3}\{0,0.1,0.1\sqrt{10},1,\sqrt{10},10\}$ \\
\addlinespace
SpecSAM-Muon
& $\rho_m$
& $10^{-2}\{0.2,0.5,1,2,5\}$ \\
& $\rho_n$
& $10^{-3}\{0,0.1,0.1\sqrt{10},1,\sqrt{10},10\}$ \\
\addlinespace
LayerSAM-SGDW
& $\rho_m$
& $10^{-2}\{0.1,0.2,0.5,1,2,5\}$ \\
& $\rho_n$
& $10^{-3}\{0,0.1,0.1\sqrt{10},1,\sqrt{10},10\}$ \\
\addlinespace
LayerSAM-Muon
& $\rho_m$
& $10^{-1}\{0.1,0.2,0.5,1,2,5\}$ \\
& $\rho_n$
& $10^{-4}\{0,0.1,0.1\sqrt{10},1,\sqrt{10},10\}$ \\
\bottomrule
\end{tabular}
\caption{Hyperparameter search grids for ResNet-50. As above, the sharpness-aware variants inherit the learning rate and weight decay from their base optimizer; only the additional perturbation-radius grids are listed for them.}

\label{tab:ResNet-50-search-grids}
\end{table}

The full search grids for the two architectures are listed in Tables~\ref{tab:vit-search-grids} and~\ref{tab:ResNet-50-search-grids}. Final configurations are selected by validation Top-1 accuracy at the reduced training budgets and are reported in Tables~\ref{tab:vit-hyperparameters} and~\ref{tab:ResNet-50-hyperparameters}.

\begin{table}[t]
\centering
\small
\setlength{\tabcolsep}{4pt}
\renewcommand{\arraystretch}{1.15}
\begin{tabular}{@{}lccccc@{}}
\toprule
Method & LR/mLR & nLR & WD & $\rho,\rho_m$ & $\rho_n$ \\
\midrule
AdamW
& $3 \times 10^{-3}$ & -- & $3 \times 10^{-1}$ & -- & -- \\
Muon
& $3\sqrt{3} \times 10^{-3}$ & $3 \times 10^{-3}$ & $3 \times 10^{-1}$ & -- & -- \\
SAM-AdamW
& $3 \times 10^{-3}$ & -- & $3 \times 10^{-1}$
& $\sqrt{21} \times 10^{-1}$ & -- \\
SAM-Muon
& $3\sqrt{3} \times 10^{-3}$ & $3 \times 10^{-3}$ & $3 \times 10^{-1}$
& $7 \times 10^{-1}$ & -- \\
SpecSAM-AdamW
& $3 \times 10^{-3}$ & -- & $3 \times 10^{-1}$
& $\tfrac{9}{7} \times 5.1 \times 10^{-4}$ & $0$ \\
SpecSAM-Muon
& $3\sqrt{3} \times 10^{-3}$ & $3 \times 10^{-3}$ & $3 \times 10^{-1}$
& $\tfrac{3\sqrt{21}}{7} \times 5.1 \times 10^{-4}$ & $\sqrt{10} \times 10^{-3}$ \\
\bottomrule
\end{tabular}
\caption{Selected hyperparameters for ViT-Small/16.}
\label{tab:vit-hyperparameters}
\end{table}
\begin{table}[t]
\centering
\small
\setlength{\tabcolsep}{4pt}
\renewcommand{\arraystretch}{1.15}
\begin{tabular}{@{}lccccc@{}}
\toprule
Method & LR/mLR & nLR & WD & $\rho$, $\rho_m$ & $\rho_n$ \\
\midrule
SGDW
& $1.6$
& --
& $10^{-3}$
& --
& -- \\
Muon
& $2 \times 10^{-2}$
& $2 \times 10^{-2}$
& $10^{-1}$
& --
& -- \\
SAM-SGDW
& $1.6$
& --
& $10^{-3}$
& $5 \times 10^{-2}$
& -- \\
SAM-Muon
& $2 \times 10^{-2}$
& $2 \times 10^{-2}$
& $10^{-1}$
& $2 \times 10^{-1}$
& -- \\
SpecSAM-SGDW
& $1.6$
& --
& $10^{-3}$
& $5 \times 10^{-4}$
& $10^{-3}$ \\
SpecSAM-Muon
& $2 \times 10^{-2}$
& $2 \times 10^{-2}$
& $10^{-1}$
& $2 \times 10^{-2}$
& $\sqrt{10} \times 10^{-4}$ \\
LayerSAM-SGDW
& $1.6$
& --
& $10^{-3}$
& $2 \times 10^{-3}$
& $\sqrt{10} \times 10^{-3}$ \\
LayerSAM-Muon
& $2 \times 10^{-2}$
& $2 \times 10^{-2}$
& $10^{-1}$
& $2 \times 10^{-2}$
& $10^{-5}$ \\
\bottomrule
\end{tabular}
\caption{Selected hyperparameters for ResNet-50.}
\label{tab:ResNet-50-hyperparameters}
\end{table}

For the ViT-Tiny/4 experiments in Appendix~\ref{sec:app:cifar-sweep}, we perform a search over learning rates \(\{0.0005,0.001\}\) and weight decay values \(\{0.02,0.05,0.1\}\) for \textsc{AdamW}. The selected learning rate and weight decay are \(0.0005\) and \(0.05\), respectively. We then use the same learning rate and weight decay for all methods when training ViT-Tiny/4.

\subsection{Datasets, models, and licenses.}\label{sec:app:licenses}
We use ImageNet-1K/ILSVRC2012 for training and validation under the ImageNet access terms, which restrict use to non-commercial research and educational purposes and note that ImageNet does not own the copyright to the underlying images. We evaluate ImageNet-ReaL annotations from the \texttt{google-research/reassessed-imagenet} repository, released under the Apache License 2.0, on the ImageNet validation images and therefore still subject to the ImageNet image terms. We evaluate ImageNet-R, whose repository is released under the MIT License. We use CIFAR-100 from the official Toronto dataset page; we could not identify a separate explicit dataset license, so we treat its license as not clearly specified. Our ViT-style backbone follows \texttt{timm} with modifications according to the Google ViT and \texttt{big\_vision} references, whose code is released under the Apache License 2.0, and our ResNet-50 implementation follows \texttt{timm}, whose code is released under the Apache License 2.0. We train models from scratch and do not use third-party pretrained checkpoints.

\subsection{Training recipe for ablation}\label{sec:app:trainingforabalation}
Here we describe the setup for the ablation experiments in
Appendix~\ref{sec:ablationfornonmatrixperturbation}. Inspired by
\citet{hassani2022escapingbigdataparadigm}, we use a lightweight
CNN--Transformer hybrid model designed for $32\times32$ CIFAR images.
The model first applies a three-layer convolutional stem, which
downsamples the input by a factor of $4$ and produces an $8\times8$
grid of $64$ tokens with embedding dimension $192$. Fixed
two-dimensional sine--cosine positional embeddings are then added,
followed by $4$ pre-norm Transformer blocks with $2$ attention heads
and MLP ratio $6$. The final prediction is obtained by global average
pooling followed by a two-layer MLP classification head.

For training, we
train for $150$ epochs on CIFAR-100 with batch size $512$, $2910$
warmup iterations, and the basic data augmentation described in
Appendix~\ref{sec:app:datasets}. For each outer optimizer, Muon or
AdamW, we select the learning rate and weight decay by grid search and
then keep them fixed when sweeping the perturbation radii. Specifically, for
the Muon outer optimizer we use learning rate $0.009$ and weight decay
$0.5$, while for the AdamW outer optimizer we use learning rate $0.006$
and weight decay $0.3$.

\clearpage
\section{Ablation experiments}
\subsection{Ablation for non-matrix parameters perturbation}\label{sec:ablationfornonmatrixperturbation}
This appendix checks whether adding a separate perturbation radius for the non-matrix parameter group is useful in a smaller CIFAR-100 setting, details of the experiments are described in Appendix~\ref{sec:app:trainingforabalation}.  For these ablations, we split the inner perturbation radius into two parts: $\rho_m$ for matrix-valued parameters and $\rho_n$ for the remaining non-matrix parameters, corresponding to the $\mW^{(\ell)}$ and $\vv$ blocks in Section~\ref{subsec:layerwise}.  Tables~\ref{tab:nonmatrix-ablation-specsam-muon} and~\ref{tab:nonmatrix-ablation-specsam-adamw} report the mean and standard deviation of validation top-1 accuracy. Overall, the results do not show a consistent improvement from taking $\rho_n>0$.

\begin{table}[h]
\centering
\small
\setlength{\tabcolsep}{5pt}
\begin{tabular}{ccccccc}
\toprule
$\rho_m$
& $\rho_n=0$
& $\rho_n=10^{-5}$
& $\rho_n=10^{-4}$
& $\rho_n=10^{-3}$
& $\rho_n=10^{-2}$
& $\rho_n=\rho_m$ \\
\midrule
$0.005$
& $73.39 \pm 0.17$
& $73.42 \pm 0.30$
& $73.20 \pm 0.26$
& $73.37 \pm 0.09$ 
& $73.34 \pm 0.15$
& $73.17 \pm 0.20$ \\

$0.02$
& $74.65 \pm 0.24$
& $74.64 \pm 0.32$
& $74.60 \pm 0.32$
& $74.68 \pm 0.38$
& $74.72 \pm 0.43$
& $74.70 \pm 0.36$ \\

$0.035$
& $74.87 \pm 0.23$
& $74.93 \pm 0.06$
& $74.94 \pm 0.17$
& $74.76 \pm 0.12$
& $74.80 \pm 0.20$
& $74.80 \pm 0.04$ \\

\bottomrule
\end{tabular}
\caption{\textsc{SpecSAM-Muon} non-matrix perturbation ablation on CIFAR-100.
$\rho_m$ denotes the perturbation radius for matrix-valued parameters, while $\rho_n$ denotes the perturbation radius for the remaining non-matrix parameters.
Each entry reports mean validation top-1 accuracy $\pm$ sample standard deviation over three random seeds.}
\label{tab:nonmatrix-ablation-specsam-muon}
\end{table}

\begin{table}[h]
\centering
\small
\setlength{\tabcolsep}{6pt}
\begin{tabular}{ccccccc}
\toprule
$\rho_m$ 
& $\rho_n=0$ 
& $\rho_n=10^{-5}$ 
& $\rho_n=10^{-4}$ 
& $\rho_n=10^{-3}$ 
& $\rho_n=10^{-2}$ 
& $\rho_n=\rho_m$ \\
\midrule
$0.0005$
& $70.62 \pm 0.50$
& $70.59 \pm 0.67$
& $70.84 \pm 0.56$
& $70.34 \pm 0.32$
& $70.48 \pm 0.46$
& $70.75 \pm 0.59$ \\

$0.001$
& $70.70 \pm 0.30$
& $71.00 \pm 0.45$
& $70.81 \pm 0.26$
& $70.70 \pm 0.46$
& $70.75 \pm 0.72$
& -- \\

$0.002$
& $70.94 \pm 0.30$
& $70.66 \pm 0.43$
& $70.92 \pm 0.19$
& $70.80 \pm 0.27$
& $70.87 \pm 0.13$
& $70.57 \pm 0.13$ \\

\bottomrule
\end{tabular}
\caption{\textsc{SpecSAM-AdamW} non-matrix perturbation ablation on CIFAR-100.
$\rho_m$ denotes the perturbation radius for matrix-valued parameters, while $\rho_n$ denotes the perturbation radius for the remaining non-matrix parameters.
Each entry reports mean validation top-1 accuracy $\pm$ sample standard deviation over three random seeds.}
\label{tab:nonmatrix-ablation-specsam-adamw}
\end{table}

\subsection{Ablation for Layerwise perturbation}\label{sec:app:layersam}

The main experiments contrast a \emph{global} Euclidean inner perturbation (\textsc{SAM}) with a \emph{layerwise spectral} one (\textsc{SpecSAM}). These two choices differ in two respects at once: relative to standard \textsc{SAM}, \textsc{SpecSAM} both (i) allocates a separate perturbation budget to each matrix-valued layer, through the $r=\infty$ block aggregation of Eq.~\eqref{eq:block-aggregation-norm}, and (ii) measures each block with the spectral norm instead of the Frobenius norm. To disentangle these two effects, we introduce \textsc{LayerSAM}, an intermediate variant that keeps the layerwise ($r=\infty$) budgeting of \textsc{SpecSAM} but replaces the per-block spectral norm with the per-block Frobenius (Euclidean) norm, i.e. $\|\cdot\|_{(\ell)}=\|\cdot\|_F$ for all $\ell\in[L-1]$. Since the Frobenius norm is self-dual with subgradient $\mG^{(\ell)}/\|\mG^{(\ell)}\|_F$, the resulting inner perturbation normalizes each matrix block independently,
\begin{equation}\label{eq:layerwise-frob-perturbation}
\vveps^*(\vtheta)
=
\left(
\frac{\mG^{(1)}(\vtheta)}{\|\mG^{(1)}(\vtheta)\|_F},
\dots,
\frac{\mG^{(L-1)}(\vtheta)}{\|\mG^{(L-1)}(\vtheta)\|_F},
\alpha_g \frac{\vg(\vtheta)}{\|\vg(\vtheta)\|_2}
\right),
\end{equation}
with the non-matrix block handled exactly as in our \textsc{SpecSAM} runs (a separate Euclidean radius $\rho_n$; see Appendices~\ref{sec:app:routing} and~\ref{sec:app:hpsearch}). \textsc{LayerSAM} therefore isolates the effect of \emph{per-layer budgeting} from that of \emph{spectral geometry}: any gap between \textsc{SAM} and \textsc{LayerSAM} is attributable to layerwise normalization, while any further gap between \textsc{LayerSAM} and \textsc{SpecSAM} is attributable to the spectral norm itself.

We evaluate both outer optimizers on ResNet-50/ImageNet-1K under the protocol of Section~\ref{sec:exp:imagenet}, with search grids and selected radii reported in Appendix~\ref{sec:app:hpsearch}. Table~\ref{tab:layersam-ablation} places \textsc{LayerSAM} alongside the corresponding non-SAM baselines and the \textsc{SAM}/\textsc{SpecSAM} variants.

\begin{table}[h]
\centering
\small
\setlength{\tabcolsep}{6pt}
\renewcommand{\arraystretch}{1.1}
\begin{tabular}{@{}llcr@{}}
\toprule
Name & Inner Step & Outer Step & Val Top-1 \\
\midrule
\textsc{SGDW}          & None                   & SGDW & $77.35_{\pm 0.07}$ \\
\textsc{SAM-SGDW}      & $\ell_2$ (global)      & SGDW & $77.99_{\pm 0.14}$ \\
\textsc{LayerSAM-SGDW} & $\ell_2$ (layerwise)   & SGDW & $77.67_{\pm 0.06}$ \\
\textsc{SpecSAM-SGDW}  & Spectral (layerwise)   & SGDW & $77.77_{\pm 0.15}$ \\
\midrule
\textsc{Muon}          & None                   & Muon & $77.05_{\pm 0.07}$ \\
\textsc{SAM-Muon}      & $\ell_2$ (global)      & Muon & $77.84_{\pm 0.03}$ \\
\textsc{LayerSAM-Muon} & $\ell_2$ (layerwise)   & Muon & $77.49_{\pm 0.04}$ \\
\textsc{SpecSAM-Muon}  & Spectral (layerwise)   & Muon & $\mathbf{78.55}_{\pm 0.06}$ \\
\bottomrule
\end{tabular}
\caption{\textsc{LayerSAM} ablation on ResNet-50/ImageNet-1K. Each block fixes the outer optimizer and varies only the inner-step geometry: no perturbation (None), a global $\ell_2$ ball (\textsc{SAM}), a layerwise $\ell_2$/Frobenius ball (\textsc{LayerSAM}, Eq.~\eqref{eq:layerwise-frob-perturbation}), and a layerwise spectral ball (\textsc{SpecSAM}). Validation Top-1 is reported as mean$_{\pm\text{std}}$ over 3 seeds; the non-LayerSAM rows are reproduced from Table~\ref{tab:main}. Bolding follows the convention of Table~\ref{tab:main}}
\label{tab:layersam-ablation}
\end{table}

Two observations stand out. First, moving from a global $\ell_2$ ball to a layerwise $\ell_2$ ball does not help: \textsc{LayerSAM} trails \textsc{SAM} with both outer optimizers ($77.67$ vs.\ $77.99$ for SGDW, and $77.49$ vs.\ $77.84$ for Muon). Per-layer budget allocation on its own is therefore not the source of \textsc{SpecSAM}'s gains. Second, the large \textsc{SpecSAM-Muon} result ($78.55$) is recovered only when the layerwise budgeting is combined with the spectral norm: replacing the spectral block norm by the Frobenius block norm (\textsc{LayerSAM-Muon}) erases the advantage, dropping the method back below even global \textsc{SAM-Muon}. With the SGDW outer step the three inner geometries remain within $\approx 0.3$ of one another, consistent with the weaker inner/outer interaction observed for non-matrix-aware optimizers in the main text. Taken together, this ablation indicates that the benefit of \textsc{SpecSAM-Muon} comes specifically from the spectral geometry of the inner perturbation, rather than from layerwise normalization alone, and that it is realized only when paired with a matrix-aware outer step. This also points to a further design for matrix-aware perturbations: coupling the per-layer budgets through a finite aggregation exponent (e.g. $r=2$) rather than the $r=\infty$ aggregation used here, which we leave to future work.

\clearpage
\section{Additional experimental results on CIFAR-100} \label{sec:app:imagenet-additional-results}

\subsection{Sharpness estimation}\label{sec:app:sharpness}
Top Hessian eigenvalues are a single scalar summary of curvature and do not
distinguish between different geometric notions of sharpness. Following
Section~\ref{subsec:geometric-sam}, we measure sharpness with respect
to the same perturbation sets that each method's inner step probes.
Fix a parameter $\vtheta\in\Theta$ and a closed perturbation set
$\mathcal{C}_\rho\subset\Theta$. Let $S$ be a fixed subset of the training
set, drawn once and reused across all methods, and let $L_S$ denote the
empirical loss restricted to $S$; we take $S$ to be the full training set.
We define
\[
  \mathcal{S}_{\mathcal{C}}(\vtheta;\rho)
  \;:=\;
  \max_{\vveps\in\mathcal{C}_\rho}
  L_S(\vtheta+\vveps)-L_S(\vtheta).
\]
We use two choices of $\mathcal{C}_\rho$, matching the two inner-step
geometries in Table~\ref{tab:method-grid}. \emph{$\ell_2$ sharpness} uses the
Euclidean ball $\{\vveps:\|\vveps\|_2\le\rho\}$ on the flattened parameter
vector. \emph{Spectral sharpness} uses a simplified version of the layerwise spectral ball: every matrix block
$\mE^{(\ell)}$ has spectral norm at most $\rho$, and the non-matrix block is
set to zero. The radii are stated in
Appendix~\ref{sec:app:sharpness-diagnostics}.
The inner maximum is approximated by projected gradient ascent with three
random restarts. For $\ell_2$ sharpness, the iterate is rescaled so its
$\ell_2$-norm does not exceed $\rho$. For spectral sharpness, each matrix
block is rescaled so its spectral norm does not exceed $\rho$, with the
spectral norm estimated by $8$ steps of power iteration, and the non-matrix
block is fixed at zero; the per-block ascent direction mirrors the (simplified)
\textsc{SpecSAM} inner step, replacing each block gradient by its
five-step Newton--Schulz approximation. These approximations introduce
small errors, so the reported values should be read as a qualitative
comparison rather than exact maxima.

\subsection{Effective-rank diagnostics}\label{sec:app:effective-rank}
For a non-zero matrix $\mA$ with non-zero singular values
$\sigma_1\ge\cdots\ge\sigma_r>0$, let $p_i=\sigma_i/\sum_{j=1}^{r}\sigma_j$.
We use the entropy-based effective rank of   \citet{roy2007effective},
\[
  \operatorname{erank}(\mA)
  \;:=\;
  \exp\!\Bigl(-\sum_{i=1}^{r} p_i\log p_i\Bigr),
\]
with $\operatorname{erank}(\mathbf{0})=0$. It lies between $1$ and
$\mathrm{rank}(\mA)$, attaining the upper bound when the non-zero singular
values are equal and decreasing as the spectrum concentrates. We apply this definition to three objects, all evaluated on the entire training dataset.

The \emph{weight effective rank} sums $\operatorname{erank}(\mW^{(\ell)})$
over the four matrix-valued parameter blocks of a transformer block,
corresponding to the QKV projection, the output projection, and the two MLP
layers. The \emph{gradient effective
rank} replaces each $\mW^{(\ell)}$ by the corresponding block gradient
$\mG^{(\ell)}(\vtheta)=\nabla_{\mW^{(\ell)}}L_S(\vtheta)$ and sums over the
same blocks. The \emph{feature effective rank} takes the matrix
$\mH\in\R^{|S|\times d}$ of block-output token-pooled features (where $d$ is
the embedding width and each row corresponds to one input in $S$),
column-centers it to obtain $\bar{\mH}$, and reports
$\operatorname{erank}(\bar{\mH}^\top\bar{\mH}/|S|)$; since this matrix is
symmetric positive semidefinite, its singular values coincide with its
eigenvalues.
For ViT-Tiny/4 we report each quantity at layers $4$ and $9$ out of $12$,
chosen as a representative early and late block, and \(S\) is taken as the full training set; the qualitative ordering
of the methods was consistent across other depths.

\subsection{CIFAR-100 radius-sweep}\label{sec:app:cifar-sweep}
The diagnostic experiments in Section~\ref{sec:exp:rho-sweep}, the rank diagnostics in Appendix~\ref{sec:app:rank-diagnostics}, and the sharpness diagnostics in Appendix~\ref{sec:app:sharpness-diagnostics} all use a single CIFAR-100 setup, so that the rank and sharpness panels can be read together as different views of the same underlying sweep. The model and data augmentation used for the diagnostic experiments are described in Appendix~\ref{sec:app:architectures} and Appendix~\ref{sec:app:datasets}. Learning rate and weight decay are selected by an AdamW grid search and reused for all four SAM-type methods, so that the sweep isolates the effect of $\rho$. For each run we record the best validation top-1 checkpoint and evaluate train top-1, the top-1 generalization gap, both sharpness measures, and the effective-rank diagnostics on that checkpoint.

\subsubsection{CIFAR-100 rank diagnostics}\label{sec:app:rank-diagnostics}
\begin{figure}[h]
\centering
\includegraphics[width=0.72\linewidth]{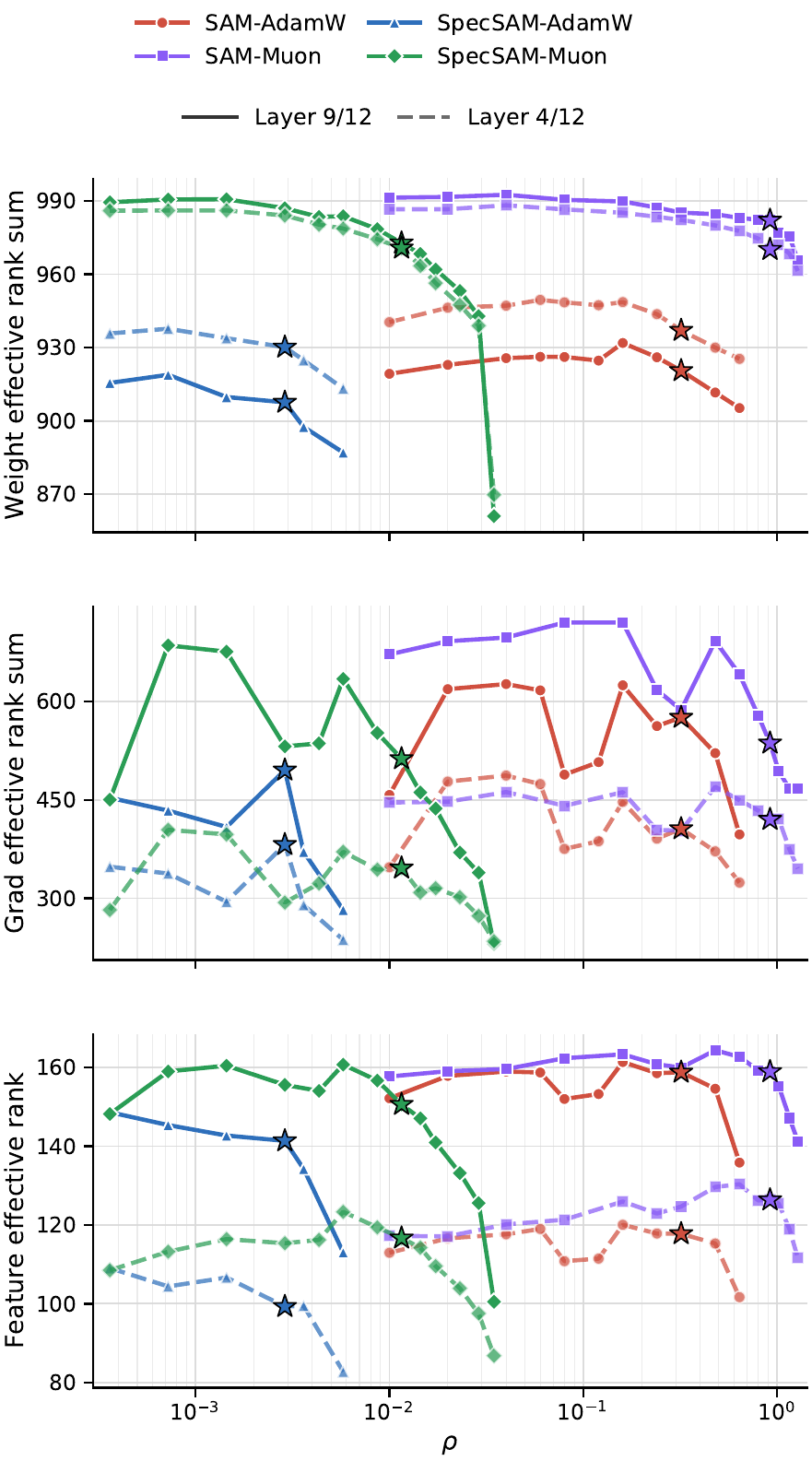}
\caption{Effective-rank diagnostics for the CIFAR-100 ViT-Tiny/4 radius sweep. Within each transformer block, weight and gradient ranks are summed over QKV, output projection, and the two MLP matrices; feature rank is the effective rank of the centered block-output covariance. Solid and dashed curves correspond to layer $9/12$ and $4/12$. Stars mark the best-validation checkpoint per method.}
\label{fig:rank-sweep}
\end{figure}

We use the CIFAR-100 sweep to ask what changes inside the network as the inner and outer geometry varies, computing the effective rank of hidden weight matrices, of loss gradients with respect to them, and of block-output feature covariances at two depths, with definitions and aggregation given in Appendix~\ref{sec:app:effective-rank}. The curves are shown in Figure~\ref{fig:rank-sweep}.

Muon-based outer updates keep the weight effective rank near the top of the range over much of the sweep and also maintain higher feature effective rank than the AdamW-based variants around the best-validation checkpoints. This may help explain why the Muon variants can profitably use larger radii: high-rank representations have more fitting capacity, which makes stronger regularization useful. However, when $\rho$ becomes too large, all three rank quantities fall and validation accuracy drops, suggesting a loss of representation capacity rather than improved generalization.

\subsubsection{CIFAR-100 sharpness diagnostics}\label{sec:app:sharpness-diagnostics}
\begin{figure}[p]
\centering
\includegraphics[width=0.95\linewidth]{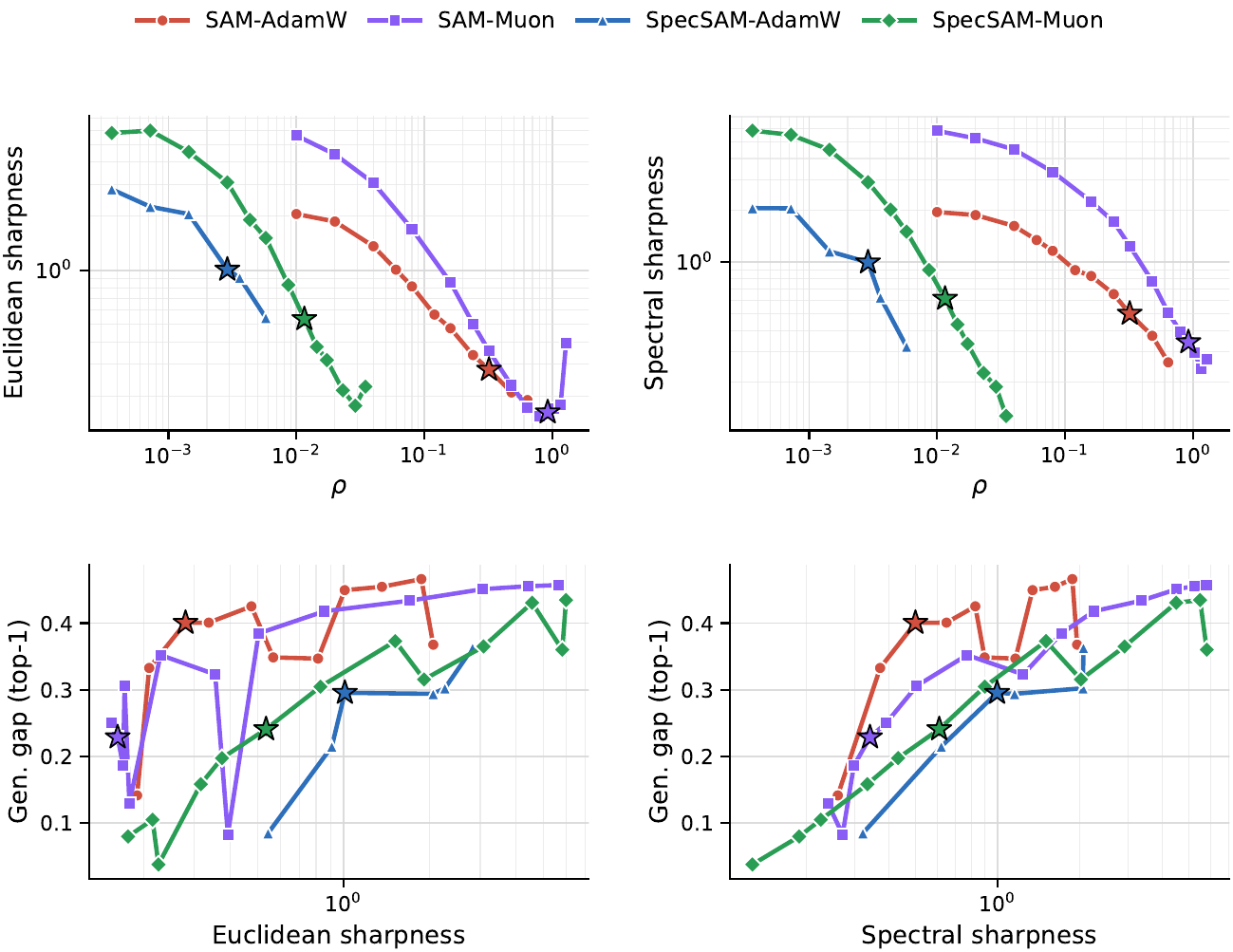}
\caption{Post-training sharpness analysis on the CIFAR-100 ViT-Tiny/4 radius sweep. The ``sharpness'' labels are described in Appendix~\ref{sec:app:sharpness}. The left and right columns use $\ell_2$ and spectral sharpness, respectively. The top row plots measured sharpness against the training radius $\rho$; the second row shows the \textbf{top-1} generalization gap against the measured sharpness. Stars mark the best-validation checkpoint per method.}
\label{fig:cifar-sharpness-frontier}
\end{figure}

On the same CIFAR-100 sweep we measure finite-radius post-training sharpness with projected gradient ascent at measurement radius $\rho=0.25$ for $\ell_2$ sharpness and $\rho=0.018$ for spectral sharpness. Figure~\ref{fig:cifar-sharpness-frontier} plots the resulting sharpness values against the perturbation radii $\rho$ in the top row, and against the top-1 generalization gap in the second row.

The top row confirms that increasing $\rho$ generally lowers both measured sharpness values, so along the sweep over $\rho$, measured sharpness and the top-1 gap are tightly entangled: larger radii flatten the loss landscape under the measurement and reduce the gap, which is useful evidence that the perturbation acts as the regularizer it is designed to be but is not a sufficient explanation of which checkpoint generalizes best. It is worth noting that, for fixed generalization gap, the \textsc{SpecSAM} variants almost always find sharper minima in terms of both \(\ell_2\) sharpness and spectral sharpness, which they penalize explicitly. The sharpness plots should also be read together with the rank plots of Appendix~\ref{sec:app:rank-diagnostics}: sharpness measures how much local sensitivity has been controlled, while rank measures how much representation capacity remains. 

\clearpage
\newpage

\end{document}